# Toward the smooth mesh climbing of a miniature robot using bioinspired soft and expandable claws

Hong Wang, Peng Liu, Phuoc Thanh Tran Ngoc, Bing Li, Yao Li, and Hirotaka Sato

*Abstract*—While most micro-robots face difficulty traveling on rugged and uneven terrain, beetles can walk smoothly on the complex substrate without slipping or getting stuck on the surface due to their stiffness-variable tarsi and expandable hooks on the tip of tarsi. In this study, we found that beetles actively bent and expanded their claws regularly to crawl freely on mesh surfaces. Inspired by the crawling mechanism of the beetles, we designed an 8-cm miniature climbing robot equipping artificial claws to open and bend in the same cyclic manner as natural beetles. The robot can climb freely with a controllable gait on the mesh surface, steep incline of the angle of 60°, and even transition surface. To our best knowledge, this is the first micro-scale robot that can climb both the mesh surface and cliffy incline.

## I. Introduction

Miniature crawling robots have been developed for off-road missions such as search and rescue, excavation, and reconnaissance missions. Their small size, lightweight, and strong navigation capabilities allow them to be deployed in complicated environments quickly. Numerous insect-scale robots have been developed with diversiform locomotion modes, including crawling [1-3], rolling [4-6], jumping[7-9], gliding [10, 11], and flying [12-14]. The actuators are diverse from traditional motors [15] and pneumatic [16] to shape memory alloy [17], piezoelectric ceramics [18], and dielectric elastomer [19]. Existing micro crawling robots, such as SMALLBug [20], 3P robot [21] DEAnsect [22], MinIAQ [23, 24], and VelociRoach [25, 26] reveal an impressive crawling ability. However, they can only locomote on a nearly level surface, which makes them unable to overcome barriers several times larger than their body size. Consequently, microrobots with climbing ability are imperative to be designed.

Only a few mini-scale robots can crawl on an incline or cliff. Li's robot [27], with three stiff hooks that served as feet, climbed an incline of 30° but at a very low speed with DEA-actuator (about 1 mm/s). Some state-of-the-art works on inclined surface climbing have focused on adapting gecko-inspired adhesion mechanisms. HAMR-E [28, 29], a quadrupedal microrobot with electroadhesion, enabled locomotion on inclines from 0° to 180°, but the stability was low on account of occasional pad detachments. Recent work reported a tethered robot [30] based on two double-pole electroadhesive pads as feet can climb on vertical flat glass at 0.75 body length per second. However, it relied too much on the materials and textures of the surface so that it cannot climb on a dusty or irregular surface, let alone a mesh surface. These robots face difficulties on some challenging terrains, such as uneven surfaces or dusty cliffy inclines owing to their requirements for smoothness of contact surfaces.

Few micro robots can travel or climb on a meshed surface because of the high risk of being stuck. When climbing on a meshed incline, a directional attachment and detachment device is required to provide impetus to move forward and disengage with the surface to prepare for the next step [31]. It is challenging to implement such a device on a micro robot. In nature, insects can manipulate their articulated legs and claws to dictate the direction of attachment and detachment. Several creatures such as cockroaches and beetles can move freely on uneven terrains, including meshed substrates owing to the tarsi and claws or pads at the tips of their legs. By cutting the tarsi and spines, cockroach (Periplaneta americana) significantly decreased their crawling speed on the meshed surface compared to the normal ones [32]. Ichikawa found that a chordal elastic organ tied to the end of the tibia of the beetle (Zophobas atratus) assisted the pulled apodeme to return smoothly. The elastic organ may help the reliable detachment of claws that engage tightly with the terrain when traversing inverted surfaces [33]. Shima et al. found that the beetle's claws (Mecynorhina torquata) remained close when swinging propodeons, and the claws opened and grasped the surface while stretching [34]. The

This work was supported by National Natural Science Foundation of China (Grant No. 51905120), Shenzhen Science and Technology Program (Grant No. RCBS20210609103901011), and Shenzhen Fundamental Research Key Program (Grant No. JCYJ20210324115812034). (Corresponding author: Yao Li, and Hirotaka Sato.)

Hong Wang is with State Key Laboratory of Robotics and System, Harbin Institute of Technology, Harbin 150001, China, with the School of Mechanical Engineering and Automation, Harbin Institute of Technology, Shenzhen 518055, China (e-mail: wanghong@stu.hit.edu.cn).

Peng Liu is with the School of Mechanical Engineering and Automation, Harbin Institute of Technology, Shenzhen 518055, China (e-mail: 19S053008@stu.hit.edu.cn).

Phuoc Thanh Tran Ngoc is School of Mechanical and Aerospace Engineering, Nanyang Technological University, Singapore, Singapore (e-mail: TRANNGOC001@e.ntu.edu.sg).

Bing Li is with State Key Laboratory of Robotics and System, Harbin Institute of Technology, Harbin 150001, China, with the School of Mechanical Engineering and Automation, Harbin Institute of Technology, Shenzhen 518055, China (e-mail: libing.sgs@hit.edu.cn).

Yao Li is with School of Mechanical Engineering and Automation, Harbin Institute of Technology, Shenzhen 518055, China (e-mail: liyao2018@hit.edu.cn).

Hirotaka Sato is School of Mechanical and Aerospace Engineering, Nanyang Technological University, Singapore, Singapore (e-mail: hirosato@ntu.edu.sg).

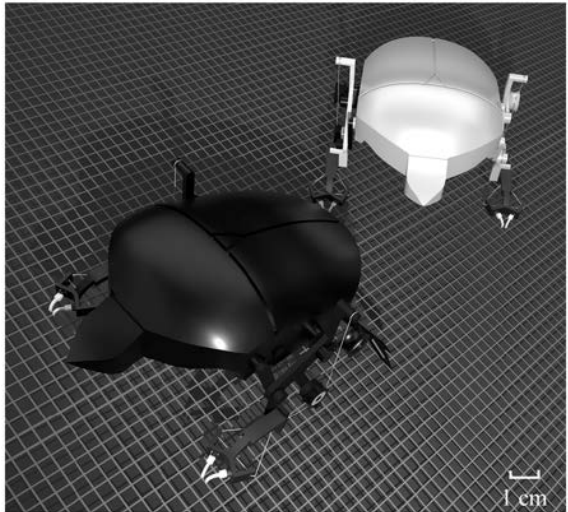

**Fig. 1** The design of a bioinspired miniature crawling robot. (a) Photo of two crawling robots on a mesh surface. (b) A beetle climbing on a mesh surface with left forearm hooked in the net and right forearm about to swing forward. (c) The whole cycle when a beetle swung its forearm forward. A rigid tarsus (c1 with claws expanded) was relaxed to be flexible (c2 flexible tarsus) and swung forward (c3 with claws closed), and lowered to subtract with expanded claws (c4 rigid tarsus).

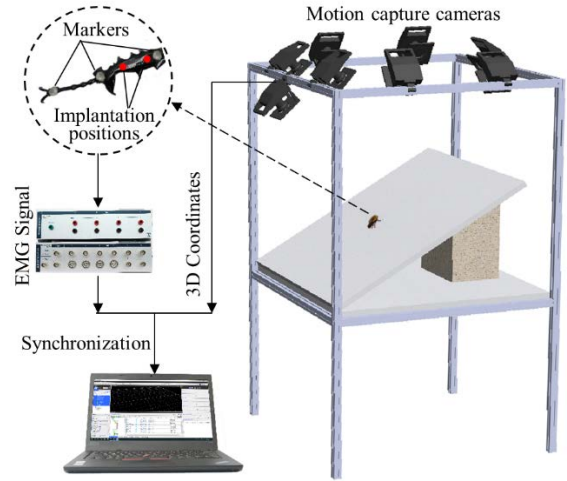

**Fig. 2** Layout diagram of the EMG signal acquisition experiment. A beetle was placed on an incline crawling freely under a motion-capture system to acquire the trajectory of the tarsus. Two silver electrodes were implanted in the tibia to collect EMG. The EMG and motion data were synchronized by a function generator.

special strategy enables the beetle to maneuver freely on the gauze or soft surface without getting stuck.

In this study, we investigated the crawling process of a beetle on an incline, and attach and detach principle of the claw. The multiple ball-joint tarsus on beetle forearm was found as a variable-stiffness structure, which became soft during swinging while rigid during stretching. A beetle-claw-inspired mesh climbing mechanism was then designed, which could bend downwards and expand the claws under the cyclic actuation of a cable-and-pully-driven system. A miniature crawling robot (Fig. 1) was assembled with the mesh climbing mechanism. The robot with the bioinspired claws walked smoothly on an inclined mesh surface. Two natural walking gaits of the beetles were imitated by the robot. The robot could maintain the phase difference between two forearms under both walking gaits while climbing on the incline under closed-loop control. Furthermore, we demonstrate the robot catching and holding the mesh incline when thrown to the substrate at a high speed. The main contribution of this work is the design and actuation of a bioinspired claw, and the development of a new strategy for micro-climbing robots to climb on the mesh surface.

## II. METHODOLOGY

### A. Study animal

Male adult flower beetle, Mecynorhina torquata (order: Coleoptera), was used in this study. The beetles were kept in separate cells of dimensions 15 cm × 10 cm × 10 cm (length × width × height) at the temperature of approximately 25°C and fed with sugar jelly every three days. The snout-vent length of the experimental animal was $55.3 \pm 2.6$ mm (mean ± s.d.), and the body mass was $5.7 \pm 1.0$ g (mean ± s.d.). Specifically, seven beetles were used in the motion capture and EMG acquisition experiments. The use of this animal has been approved by the Animal Ethical and Welfare Committee (IACUC-2020026).

### B. Motion capture and EMG acquisition during crawling

To study the bending time of beetle tarsus on an incline, two-wire electrodes (silver: 76.2 μm uncoated diameter, 139.7 μm coated diameter; A-M Systems, Sequim, WA, USA) were implanted into the muscle tissues in the beetle tibia (Fig. 2). The outer part of the electrode was glued on the tibia to avoid joggling while crawling. Beetles were anesthetized in a sealed bag filled with carbon dioxide before being implanted. Meanwhile, three 3-mm diameter retro-reflective markers were pasted on the tibia's tip for motion tracking.

After electrode implantation and marker fixation, the beetle was placed in a small box for ten minutes to rejuvenate. Afterwhile, the beetle was placed under the motion-capture system (V5, Vicon Motion Systems Ltd., Oxford, UK; 500 fps) to walk freely on an incline. A wood stick was used to tap the rear side of the beetle to initiate the forward walking. While Vicon recorded the tarsus's position, the EMG signal was collected by bioelectricity acquisition equipment (PowerLab PL3508, AD Instruments Pty Ltd Bella Vista, Australia) with a sampling rate of 1000 Hz. A function generator synchronized the motion-capture system and EMG acquisition equipment to align the 3D coordinates of the leg marker and the EMG spikes of the tarsus muscle.

### C. Structure design of a bio-inspired crawling robot

The overall size of the robot is 80 mm long and 70 mm wide, equipped with two 20-mm tarsi and 4.5-mm claws. The control

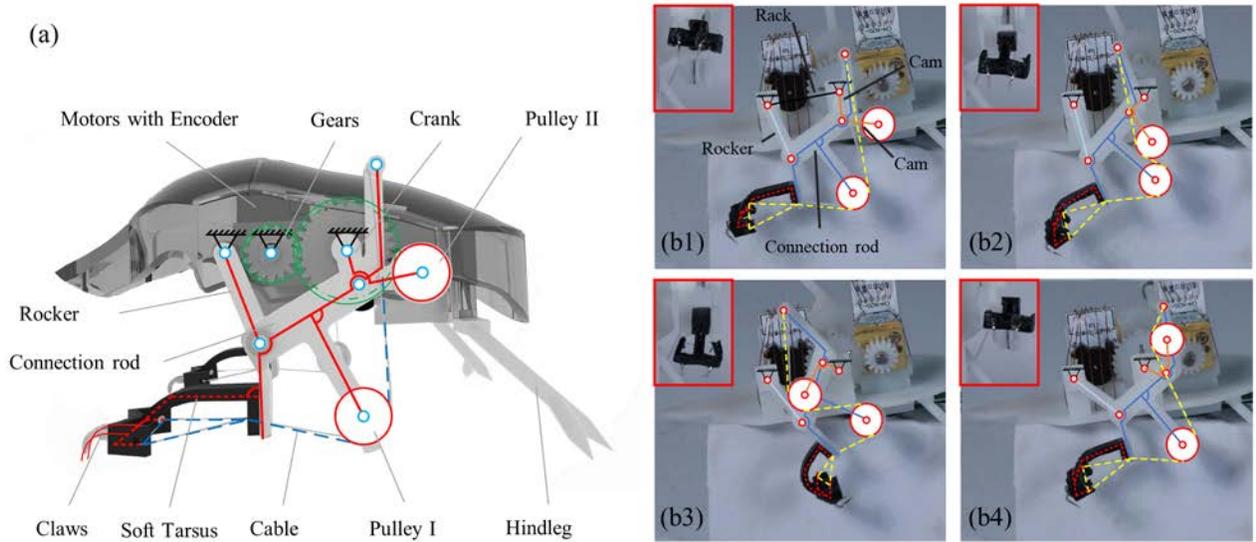

**Fig. 3** (a) Schematic diagram of transmission mechanism, a four-bar mechanism and rope-driving mechanism. A cam fixed on the crank was used to drive the four-bar mechanism (red solid lines), and the pullies drove the rope (blue dashed line) to deform the claw (red dashed line) periodically. (b) Side views of the bioinspired claw during a crawling cycle. The cam periodically rotated so that the rope was pulled to drive the tarsus bending and claws expanding.

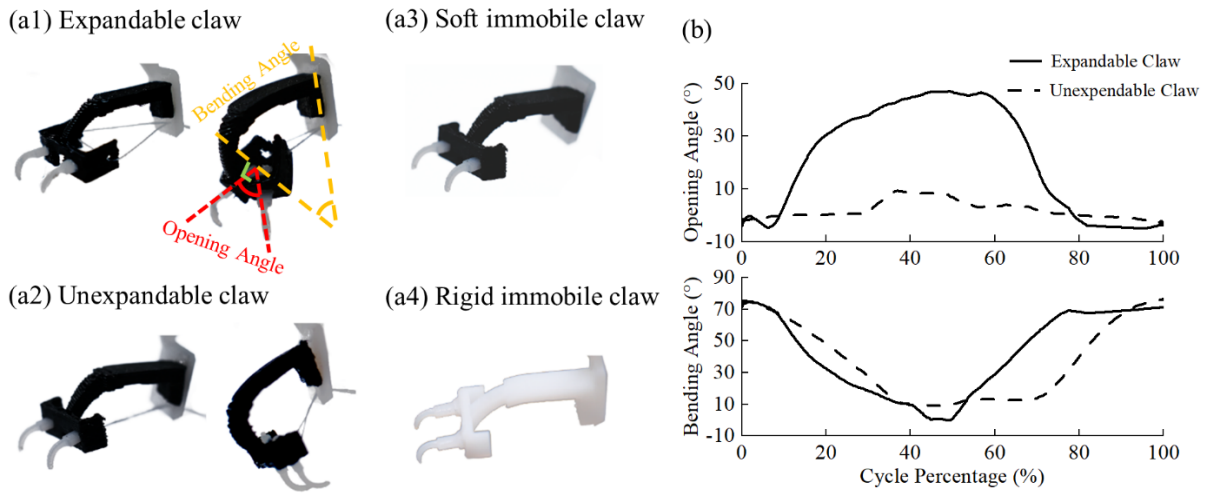

**Fig. 4** (a) The structure of the four claws. The rest (left) and bending (right) status for the expandable (a1) and unexpendable (a3) claws. The images of the soft immobile claw (a2) and rigid immobile claw (a4). (b) The change of opening angle and bending angle for expandable claws and unexpendable claws in a crawling cycle.

board, power board, and batteries have a mass of 1 g, 2 g, and 1.5 g, respectively. Mass distribution is shown in Table I, and the total mass of the robot is 50 g. The mechanical structures were manufactured using 3D printing.

Most legged robots crawl via multi-link mechanisms, which are unable to help robots climb on a meshed incline. While beetles are crawling, the tarsi, driven only by a single claw retractor muscle within the tibia [35], would bend and become stiff and the claws would open before grasping the substrate, which helps with the meshed surface crawling [31, 34]. Inspired by the strategy, we combined a four-bar mechanism and a cable-and-pully-driven mechanism as the crawling leg to replicate the function of a real forearm of beetles (Fig. 3(a)).

A pully (Pully I) was equipped on the bottom of the connecting rod to wind the cable. The cable passed through two holes on the tarsus and a hole on the front side of the connection rod and was finally fixed on the top of the connection rod. Another pully (Pully II) was installed on a cam, which was fixed with the crank. As the crank rotated, the cam would pull the cable back and forth. Thus, the artificial tarsus bent, and the claw parts opened periodically, identical to the natural claws of the beetle. The timing of tarsus bending was determined by the angle between the cam and the crank, which was inspired by the beetle crawling on the 30-degree incline. A cycle of the four-bar mechanism and the cable-and-pully-driven mechanism is shown in Fig. 3(b). The main parameters of the four-bar mechanism are listed in table II.

Table I  Mass Distribution of Robot

| Component | Mass (g) |
|---|---|
| Control Board | 1 |
| Power Board | 2 |
| Actuators | 8.5 × 2 = 17 |
| Chassis | 30 |
| Total | 50 |

Table II Parameter of Four-bar Mechanism

| Structure | Length (mm) |
|---|---|
| Crank | 5 |
| Connection rod | 15 |
| Rocker | 13 |
| Rack | 17 |
| Cam | 11 |

Four different claws were designed for the crawling test (Fig. 4(a)). The expandable claw, with two holes on both sides of the tarsus to thread the cable, can periodically bend and open the claw parts following the cam, while an unexpandable claw can only bend the tarsus without opening the claw parts. In contrast, the soft claws (*Thermoplastic polyurethanes*) and rigid immobile claws (*Photosensitive resin*) cannot bend automatically in the cycle.

*D. Actuation and control system*

Two geared DC motors (GA12-N20, 16 × 12 × 10 $mm^3$) equipped with hall encoders at the end were chosen as the primary power actuators, which were placed in the back of the robot. DC motors still reveal advantages in power-dense, travel stroke, and output torque compared to the miniature actuators such as shape memory alloy, piezoelectric actuators and artificial muscle. Owing to the requirements of high tensile force to bend the claws and move on the incline, a large gear reduction of 150:1 was employed to apply enough torque, which is about 0.02 N·m. Two additional pairs of gears were used to connect the motors and the driving mechanisms. The motors were deployed parallelly to actuate both legs respectively so that two forearms can be controlled independently. The working voltage of the motors was 5 V. The rotational speeds of the legs were measured using a Hall encoder directly fitted to the motor's shaft (24 counts per cycle).

Two single-cell 3.7 V, 150 mAh LiPo batteries were used to power the microcontroller, the motor driver, the sensors, and the DC motors. A tiny PCB board (33 mm × 30 mm, Fig. 5(a)) was custom-designed to achieve crawling control, which contains a microcontroller ATmega 2560 and a motor driver TB6612FNG. The structure of the whole control system is

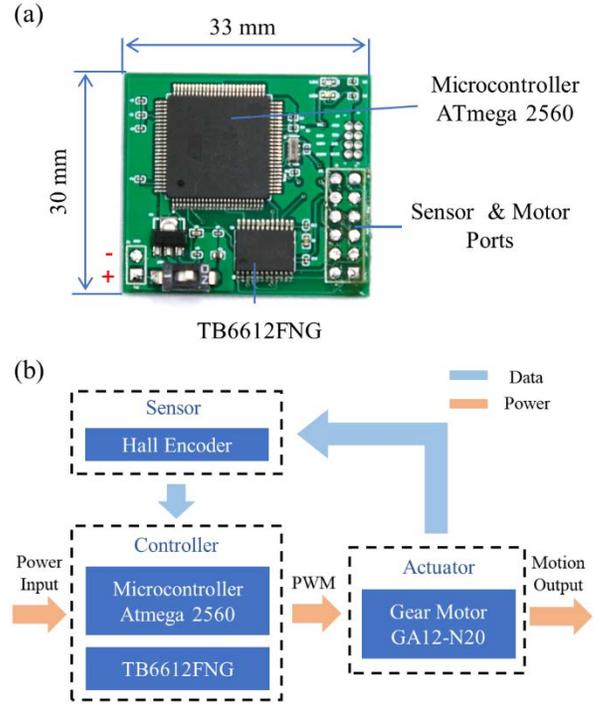

**Fig. 5** (a) The photo of the custom-designed PCB board. (b) Block diagram of the control system, with blue arrow as data transmit and orange arrow as power transmit.

shown in Fig. 5(b). Closed-loop PID control was used to coordinate the speeds of two motors and maintain the walking gait of the robot.

*E. Robot system performance*

In comparing the effect of four artificial claws (expandable claw, unexpandable claw, soft immobile claw, and rigid immobile claw), the robots equipped with the claws were placed on a meshed incline (square mesh: 2.56 × 2.56 $mm^2$). The Vicon motion capture system was used to record the trajectories of the claws and the robot. There were three markers on the robot, one on the torso and two on both forearms, respectively.

Success rate and crawl speed were chosen to estimate the performance of robots on a meshed incline. The climbing success rate was calculated based on the probability of claw stuck. The robot was supposed to crawl continuously for 9 steps under the focal range of the Vicon system. The number of steps performed before the stuck was recorded in each trial. The trial was repeated 9 times for each claw. Then the success rate equaled to the sum of steps before stuck divided by the total recorded steps, e.g., when a robot crawls 4 steps smoothly and gets stuck at the fifth step, the success rate is 80%. The climb speed was the average speed from the initial to the stuck position.

As for the close-loop controlled climbing experiment on the incline, the robot was controlled to crawl at two different gaits, gallop and tripod gaits, whose phase differences between the two forearms were approximately 0° and 180°, respectively. In the mesh climbing trials, the robot crawled on an incline of 30°, and the phase difference changes were recorded. In comparing the open-loop and the closed-loop controlled trials, the initial phase difference between the two forearms was approximately 180°. The phase difference change during 7 consecutive steps was computed for each trial. Meanwhile, the crawling speeds of the close-loop controlled trials under two different gaits were computed to evaluate the effects of the gaits. All experiment was conducted under the motion capture system and the substrate was foam board.

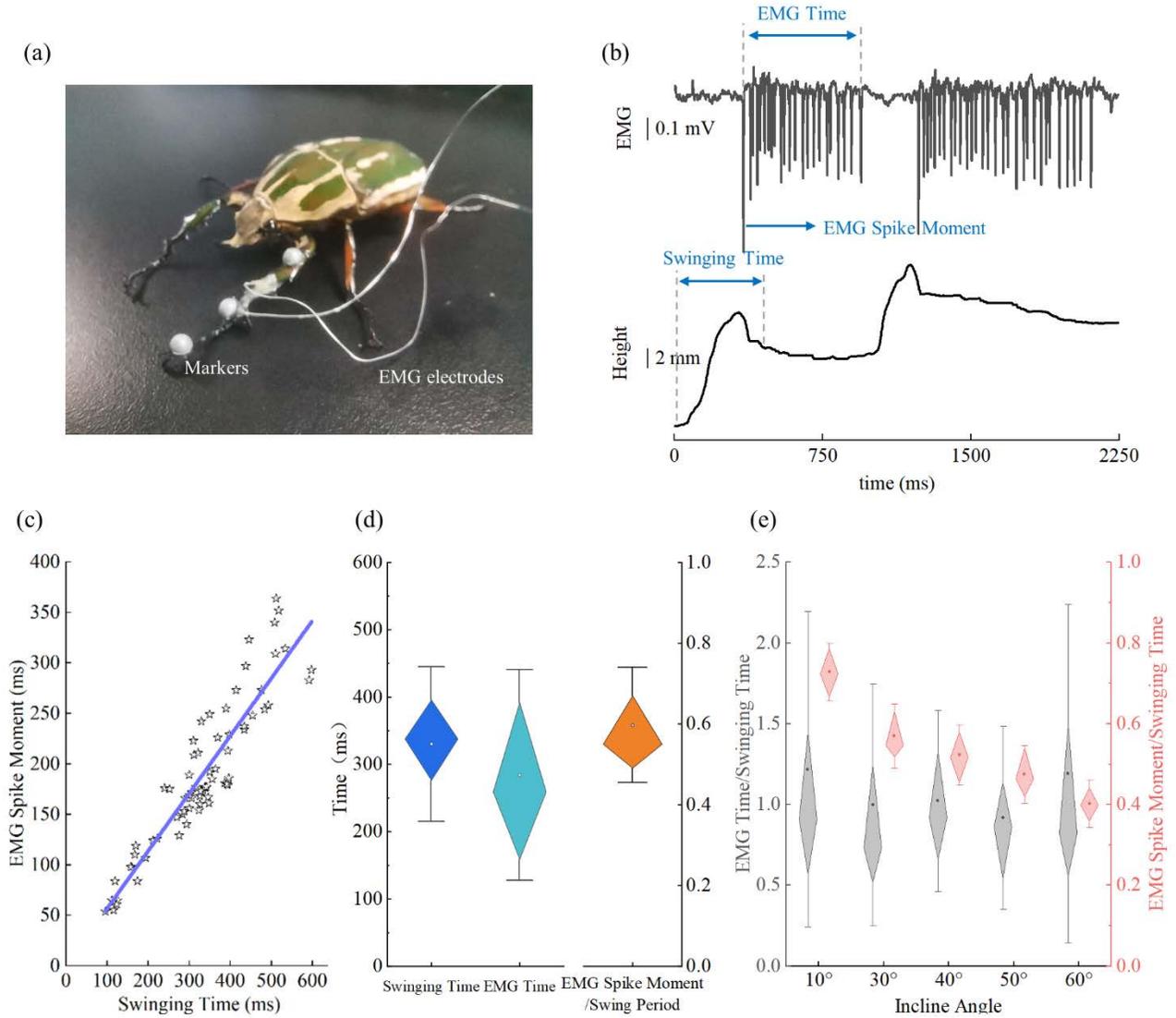

**Fig. 6** The analyses on the tarsus motion and the EMG of tarsus muscle. (a) The left tarsus of a beetle was implanted with two extrudes, and three markers were attached on left forearm for motion capture. (b) Representative EMG signal synchronized with tarsus height during two crawling periods. The tarsus height increased with the climbing steps. EMG time is the interval between the first spike to the last spike among the consecutive spikes, and EMG spike moment is the moment when the first spike emerged. Swinging time is the interval when tarsus is in the air, which begins at the vertical velocity of tarsus larger than 40 mm/s and ends at the velocity smaller than 10 mm/s. (c) The relationship between swinging time and EMG moment in crawling on a 30° incline, with star dots as the experimental results and blue line as fit curve. (d) The box plot of swinging time, EMG time, and the ratio of EMG spike moment to swinging time in crawling on a 30° incline. (e) The ratio of EMG time to swinging time and EMG spike moment over swinging time on the incline of different angles.

## III. RESULT AND DISCUSSION

*A. The analysis of tarsus motions in beetle's climbing*

The EMG signal and the tarsus height were synchronized as in Fig. 6(b). To normalize the EMG time and EMG moment among different steps, we define EMG time-ratio as the ratio of EMG time to swing time, and EMG moment-ratio as the ratio of EMG moment to swing time. Fig. 6(c) illustrates EMG moment is increasing with swing time (Pearson coefficient is 0.88, $p<0.05$) on the incline of 30° (N = 5, n = 342), and EMG moment-ratio is nearly constant, which is about $0.60 \pm 0.14$ on

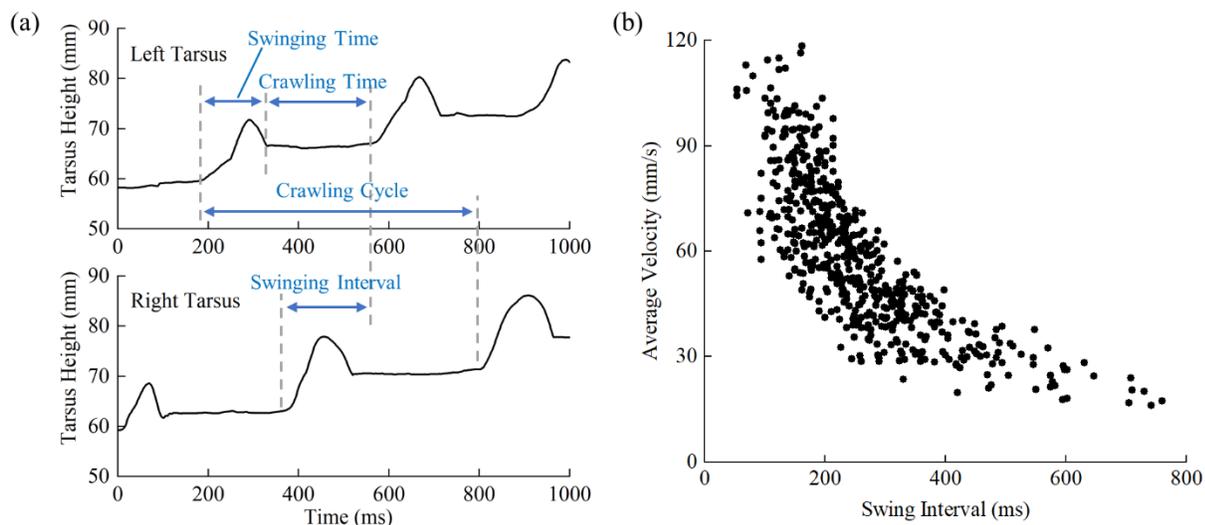

**Fig. 7** (a) Trajectories of left and right tarsus while a beetle was crawling. A single crawling period for one leg was divided into two stages: swinging time and crawling time. The period started with left tarsus lift and ended with right tarsus about to lift for a next step was defined as a crawling cycle. Swinging interval is defined as the period from the start of right tarsus swinging to the start of left tarsus swinging. The average velocity is defined as the distance beetle traveled in the cycle divides by the crawling cycle time. (b) The relationship between the average velocity and the swinging interval follows a tangent function.

the incline of 30°, as shown in Fig. 6(d) (N = 7, n = 94). Thus, tarsus bending is a controlled behavior according to the leg swing time. Beetles bend their tarsus at approximately fixed timing of swing time on the same incline.

While the angle of the incline changes, the EMG moment-ratio changes as well. Fig. 6(e) shows that when the incline angle increases, the EMG moment-ratio decreases, while the EMG time remains unaffected. That means when the beetle comes to a steeper incline, beetle tends to bend their tarsus earlier.

Philipp performed electromyographic recordings while the insects (*Carausius morosus and Cuniculina impigra*) stood on the surface under different situations and on the surface of a platform in horizontal, vertical or inverted positions, as well as during rotations of the platform. Higher frequencies of tonic units were detected during rotations compared with the stationary phases [36], meaning the insects could detect the angle changing of the surface. In our experiments, beetles bent the tarsus earlier to attach to the surface while the incline was steeper. It might be because beetles can sense the different inclines and attach to the surface in advance to avoid toggling or falling from the incline.

*B. The walking gait analysis in beetle's climbing*

Most insect species crawl with variations of the alternating tripod gait. While swinging one triangle of legs forward, they held the opposite triangle stable [37]. Researchers also observed synchronous (in-phase) stepping of a leg pair in some specific situations, when crossing an obstacle or for a few steps when the animal started to walk [38]. The additional gallop-like gait was also discovered in beetles [39]. Normally, beetles crawl with the tripod gait (asynchronous gait), in which one forearm swings earlier than the other one. While beetles tend to take a gallop gait (synchronous gait) when a fast escape response is required, in which both forearms swing together [40].

The height difference of both tarsi recorded by the motion capture system is shown in Fig. 7(a). The relationship between average velocity and swinging interval is shown in Fig. 7(b). The results reveal that the average velocity over the crawling cycle decreased as the swinging interval increased. The general trend was approximately a tangent function (p<0.05). When the swinging interval decreased to zero, which meant two forearms acted synchronously, the beetle moved faster. Another insect species (*Petrobius brevistylis*) shared the gallop gait in primitive jumping but alternated in metachronal order: back, middle, and front, which was found highly related to fast escape jump reactions [41]. In our experiments, the results also show that beetles using the gallop gait moved faster than the tripod gait. We hypothesized that the gallop gait can initiate beetles to escape from danger or predators more rapidly.

*C. Bio-inspired design of a climbing robot*

The claw motion strategy found by Shima [33] and Thanh [31] revealed that beetles closed the claws and softened the tarsus while raising and swinging their legs forward, whereas beetles expanded the claws and hardened the tarsus while depressing their legs to the surface. Inspired by the strategy, a combination of a four-bar mechanism and a cable-and-pully-driven mechanism was designed to mimic the motions of beetle tarsi and claws. Hence, the robot can perform periodic tarsus bending, claws opening and crawling with only one motor, which greatly reduces the energy consumption and the number of actuators, and the simplify the control system [42].

While a beetle was crawling upward on an incline, the claws on the mid and hindlegs remained close during the whole crawling cycle. The rear legs primarily acted for supporting the torso to avoid falling. Spagna found that leg spines on spiders increased the mobility on nets, which resulted in more effective distributed contact with asperities. The ghost crabs without natural leg spines increased mobility on wire mesh with additional artificial spines [31]. Tibial spurs of beetles were also found to help generate propulsion in narrowed spaces [43]. Therefore, a pair of tibial spur-like legs was designed to support the robot while climbing on a meshed incline.

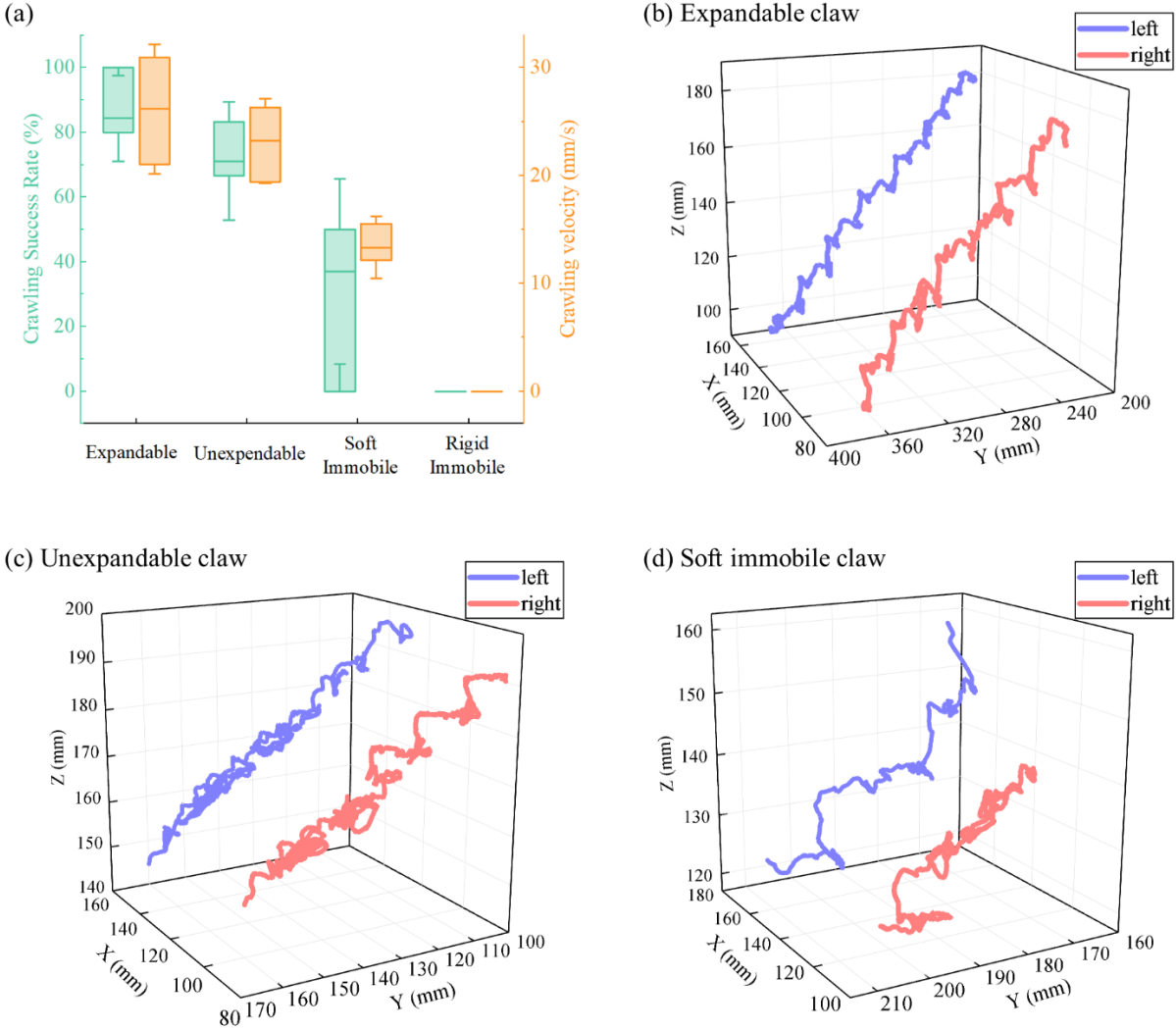

**Fig. 8** The comparison among the crawling performances of the four claws. (a) Crawling success rate and crawling velocity of a robot equipped with the four claws. Expandable claws gave the best performance, while the immobile claws cannot climb smoothly. (b-d) Trajectories of the left and right tarsus of the robot while climbing. (b) The trajectory of the tarsus with expandable claw was smooth. (c) The trajectory of the tarsus with unexpandable claw was overlapped. (d) The trajectory of the tarsus with soft immobile claw was messy, and the claw stuck in the mesh frequently.

*D. Comparison of four different kinds of robot claws*

Shima has conducted a series of experiments to verify the function of the open-close mechanism of beetle claw and applied it to a vehicle on a flat surface covered by a mesh of holes [34]. Nevertheless, they ignored the periodic variation of tarsus stiffness, which might play an important role in crawling on the mesh.

To verify the natural walking strategy of beetle, four different kinds of claws (expandable claw, unexpandable claw, soft immobile claw, and rigid immobile claw) were tested on a 30-degree mesh incline. The opening angle and bending angle in a cycle for expandable claw and unexpandable claw are illustrated in Fig. 4(b). The shape of the unexpandable claw remained unchanged while the opening angle of the expandable claw opened periodically from -5° to 45° with the rotating of the crank.

Among four kinds of claws, the expandable claw revealed the best performance, with a success rate of 84.3% ± 13.3% (n=9) and a climbing velocity of 26.18 ± 6.00 mm/s (n=9, Fig. 8(a) ). The performance of the unexpandable claw was slightly worse than the expandable claw, 71.2% ± 18.3% in success rate and 23.23 ± 3.92 mm/s in velocity. The soft claw showed a much worse performance in climbing, with a success rate of only 37.0% ± 28.6% and a climbing velocity of 13.34 ± 2.86 mm/s. The rigid claw did not support climbing on the mesh (Supplementary Movie 1).

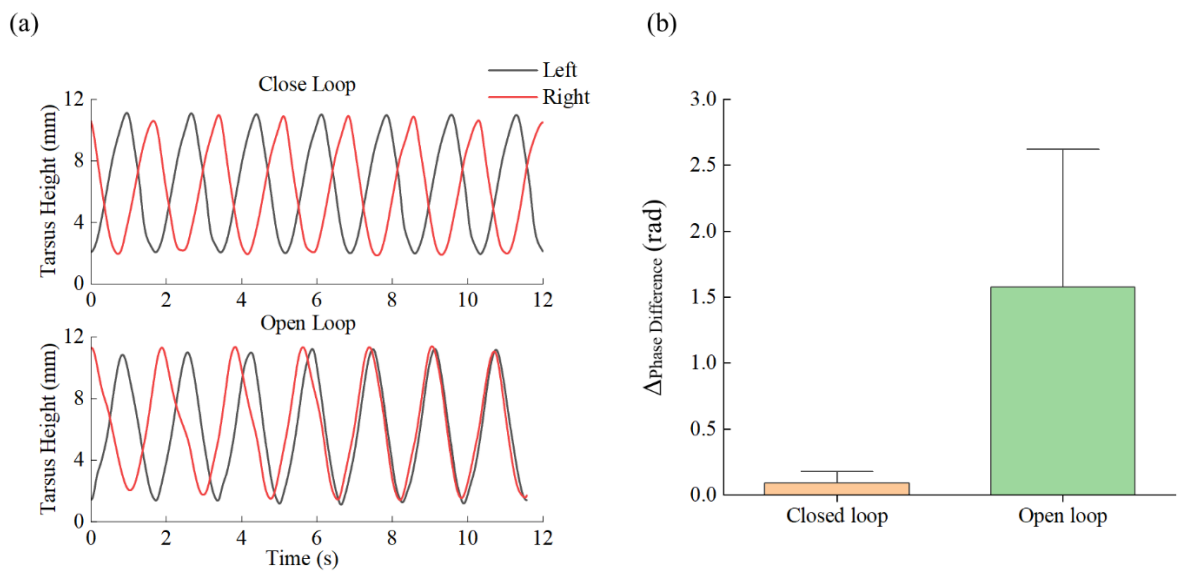

**Fig. 9** The comparison between open-loop and closed-loop control. (a) Time series of the phase changing of the left and right claws under open-loop and closed-loop control. Gray line indicates the left tarsus and red line indicates the right tarsus. (b) The change of the phase difference between two claws under closed-loop and open-loop control. The initial phase difference between two claws was approximately 180°. The change of phase difference is defined as the change from initial phase difference to the phase difference of the seventh step.

The trajectories of the expandable claw, unexpandable claw, and soft immobile claw were shown in Fig. 8(b-d), respectively. The expandable claw showed a smooth track. The trajectory of the tarsus with unexpandable claws was overlapped. The trajectory of the tarsus with soft immobile claws was messy, and the claw was stuck in the mesh frequently, due to the unreliable detachment.

The tarsus of beetle is a ball-socket structure [31]. When the tendon within the tibia is pulled, the tarsus bends downwards and turns rigid and the claws open. While the tarsus returns to its original position and turns soft and claws close as the tendon is released. Beetles use the strategy to attach and retract securely on meshed surfaces. In our experiment, the unexpandable claw only imitates the function of tarsus with claws open constantly while the expendable claw replicates both tarsus and claws. Our result reveals that periodic variation of tarsus stiffness is more significant for beetle climbing on a mesh, while the open-close mechanism of the claw part is not always necessary in mesh climbing for beetles.

*E. The walking gait control of the climbing robot*

Beetles bent their tarsi and expanded claws while lowering the forearm to the substrate for safe attachment. the robot mimicked the strategy to achieve mesh climbing. A PID controller was used in a closed-loop system with hall encoders to keep a stable walking gait between two forearms (Supplementary Movie 2). Fig. 9(b) presented the change of phase difference after 7 consecutive steps (n=5). The open-loop control yielded an apparent change of phase difference while the change of phase difference in the closed-loop was not remarkable, which means the PID controller can stabilize the gait of the robot.

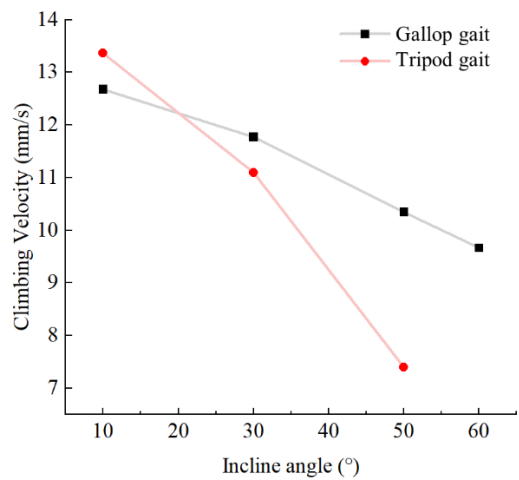

**Fig. 10** The velocity comparison between two walking gaits of the robot under different incline angles. The red line indicates the tripod gait, and gray line indicates the gallop gait.

Beetles adopted two gaits in walking, the gallop gait and the tripod gait. With the help of the closed-loop control, the robot can crawl with either gait. The robot was placed on the inclines of 10°, 30°, and 50°. The average velocities in a 15-s period for both gaits were presented in Fig. 10 (n = 5).

The result shows that as the incline angle increases, the average velocity of the robot decreases for both gaits. Since the stiffness of the claw was constant, a steeper incline caused a greater load for the robot, resulting in a smaller bending angle of the claw at each step, and hence a smaller stroke of each step. All tests were conducted under the same rotating speed of the motor, so smaller strokes of each step resulted in smaller crawling velocity. When the robot crawled with the tripod gait, there was only one tarsus to support the weight of the robot, while two tarsi to support the weight in the gallop gait. As a result, the stroke with tripod gait was smaller than that with gallop gait. Thus, the velocity of tripod gait decreased faster than that of gallop gait with the increase of incline angle. Moreover, the robot with a gallop gait can reach 10 mm/s on an incline of 60°. In comparison, the robot can only climb on a 50° incline with approximately 7.5 mm/s with a tripod gait. The robot can also climb on a 30-degree mesh surface with a velocity of 10 mm/s via gallop gait and 8 mm/s via tripod gait, respectively (Supplementary Movie 3).

## V. CONCLUSION AND FUTURE WORK

The analysis of the EMG signals of the beetle's tarsus muscle revealed the law of tarsus motions in climbing. The beetles increased the stiffness of the tarsus and expanded the claw while attached to an inclined surface. Furthermore, the timing of claw motion appeared earlier on a steeper incline to increase the attaching reliability. Then a bio-inspired claw mechanism was designed and actuated to imitate the motions of a beetle claw, whose artificial claws can bend and expand periodically. The bio-inspired claw revealed the best climbing performance on the mesh surface compared with other claws that cannot bend and expand. Then an insect-scale climbing robot was designed to crawl smoothly on both meshed surfaces and steep inclines. Furthermore, two walking gaits of the beetles were reproduced by the robot in climbing. The walking gait was closely relative to the average velocity in crawling. Under a closed-loop control, we demonstrated that the gallop gait was more efficient in climbing on a steeper surface. The proposed crawling mechanism revealed a good mesh surface adhesion ability and enabled the robot to grasp the substrate even under high-speed impact, which shows the potential application in aircraft mesh surface landing (Supplementary Movie 4).

Next, we would like to study the steering strategy of beetles while climbing on an incline, especially the bilateral asymmetrical motions of both forearms in stretching and swinging. Then we would apply the strategy to the climbing robot. A gyro sensor would be added onboard to control the turning angle precisely. Furthermore, we would explore the possibility of equipping four claws on four legs to increase the freedom of crawling, such as backward and sideward crawling. We would like to see the robot crawling on a vertical mesh surface like an actual insect in the future.

## APPENDIX

Movie S1. Performance of robots with different claws on the meshed surface.

Movie S2. Comparison between open-loop control and closed-loop control.

Movie S3. Demonstration of the robot on the meshed surface and steep incline.

Movie S4. Trials of robot catching mesh surface under high-speed impact.

## ACKNOWLEDGMENT

This work was supported by National Natural Science Foundation of China (Grant No. 51905120), Shenzhen Science and Technology Program (Grant No. RCBS20210609103901011), and Shenzhen Fundamental Research Key Program (Grant No. JCYJ20210324115812034). The authors would like to thank Mr. Ma Songsong, Mr. Yan Ze, Mr. Li Keyu, and Mr. Tang Lingqi for their assistance in organizing experimental materials and maintaining experimental facilities.